%% file: main.tex
\documentclass{article} 

\usepackage[accepted]{icml2025}

\input{math_commands.tex}

\usepackage{hyperref}
\usepackage{url}
\usepackage{enumitem} 
\usepackage{booktabs}
\usepackage{graphicx}
\usepackage{subcaption}
\usepackage{natbib}
\usepackage{multirow}

\title{What Do VLAs Actually Inherit from VLMs?}

\author{Chuheng Zhang \\
Microsoft Research \\
Corresponding to \texttt{zhangchuheng123@live.com}
}

\icmltitlerunning{How Do VLAs Effectively Inherit from VLMs?}

\begin{document}

\twocolumn[
\icmltitle{How Do VLAs Effectively Inherit from VLMs?}
\icmlsetsymbol{equal}{*}

\begin{icmlauthorlist}
\icmlauthor{Chuheng Zhang}{msr}
\icmlauthor{Rushuai Yang}{hkust}
\icmlauthor{Xiaoyu Chen}{thu}
\icmlauthor{Kaixin Wang}{msr}
\icmlauthor{Li Zhao}{msr}
\icmlauthor{Yi Chen}{hkust}
\icmlauthor{Jiang Bian}{msr}
\end{icmlauthorlist}

\icmlaffiliation{msr}{Microsoft Research}
\icmlaffiliation{hkust}{The Hong Kong University of Science and Technology}
\icmlaffiliation{thu}{Tsinghua University}

\icmlcorrespondingauthor{Chuheng Zhang}{chuhengzhang@microsoft.com}

\icmlkeywords{VLA, Robotics}

\vskip 0.3in
]



\printAffiliationsAndNotice{}  

\input{sections/abstract}
\input{sections/introduction}
\input{sections/related_work}

\input{sections/methods}
\input{sections/experiments}
\input{sections/conclusion}

\bibliography{iclr2026_conference}
\bibliographystyle{iclr2026_conference}

\clearpage
\onecolumn
\appendix
\input{sections/app_benchmark_codebase}
\input{sections/app_std}

\end{document}

%% file: math_commands.tex
\usepackage{amsmath,amsfonts,bm}

\def\eqref#1{equation~\ref{#1}}

\def\1{\bm{1}}

\DeclareMathAlphabet{\mathsfit}{\encodingdefault}{\sfdefault}{m}{sl}
\SetMathAlphabet{\mathsfit}{bold}{\encodingdefault}{\sfdefault}{bx}{n}



%% file: sections/abstract.tex
\begin{abstract}
Vision-language-action (VLA) models hold the promise to attain generalizable embodied control.
To achieve this, a pervasive paradigm is to leverage the rich vision-semantic priors of large vision-language models (VLMs).
However, the fundamental question persists: \emph{How do VLAs effectively inherit the prior knowledge from VLMs?}
To address this critical question, we introduce a diagnostic benchmark, GrinningFace, an emoji tabletop manipulation task where the robot arm is asked to place objects onto printed emojis corresponding to language instructions. 
This task design is particularly revealing -- 
knowledge associated with emojis is ubiquitous in Internet-scale datasets used for VLM pre-training, yet emojis themselves are largely absent from standard robotics datasets. 
Consequently, they provide a clean proxy: successful task completion indicates effective transfer of VLM priors to embodied control.
We implement this diagnostic task in both simulated environment and a real robot, and compare various promising techniques for knowledge transfer.
Specifically, we investigate the effects of parameter-efficient fine-tuning, VLM freezing, co-training, predicting discretized actions, and predicting latent actions. 
Through systematic evaluation, our work not only demonstrates the critical importance of preserving VLM priors for the generalization of VLA but also establishes guidelines for future research in developing truly generalizable embodied AI systems.
\end{abstract}


%% file: sections/introduction.tex
\section{Introduction}
\label{sec:introduction}

One central mission of the embodied intelligence community is the pursuit of generalist agents for real-world robotic control at scale. 
While early generalist agents~\citep{reed2022generalist,brohan2022rt1} were trained from scratch, training vision-language-action models (VLAs) based on large vision-language models (VLMs) has emerged as the dominant paradigm due to the rich visual-semantic priors of VLMs. 
The fundamental promise of this approach lies in knowledge transfer -- leveraging the rich open-vocabulary visual recognition and semantic reasoning capabilities of pre-existing VLMs~\citep{liu2023visual,achiam2023gpt,team2023gemini,bai2025qwen2,wang2025internvl3}, acquired through Internet-scale pretraining, for direct application to action prediction. 
Recent VLAs developed under this paradigm have demonstrated remarkable capabilities across diverse manipulation tasks, such as RT-2~\citep{zitkovich2023rt2}, OpenVLA~\citep{kim2024openvla}, Octo~\citep{mees2024octo}, $\pi_0$~\citep{black2024pi_0}, and GEN-0~\citep{generalist2025gen0}.

Despite the clear trend of training VLAs based on VLMs, fundamental questions persist:
\begin{itemize}[leftmargin=*]
    \item What is the minimal diagnostic probe to benchmark how well VLAs inherit VLM priors?
    \item To what extent do VLMs genuinely contribute to VLA generalization capabilities?
    \item What constitutes the best practice for VLA pre-training and fine-tuning to effectively inherit VLM priors?
\end{itemize}

\begin{figure*}
    \centering
    \includegraphics[width=0.9\linewidth]{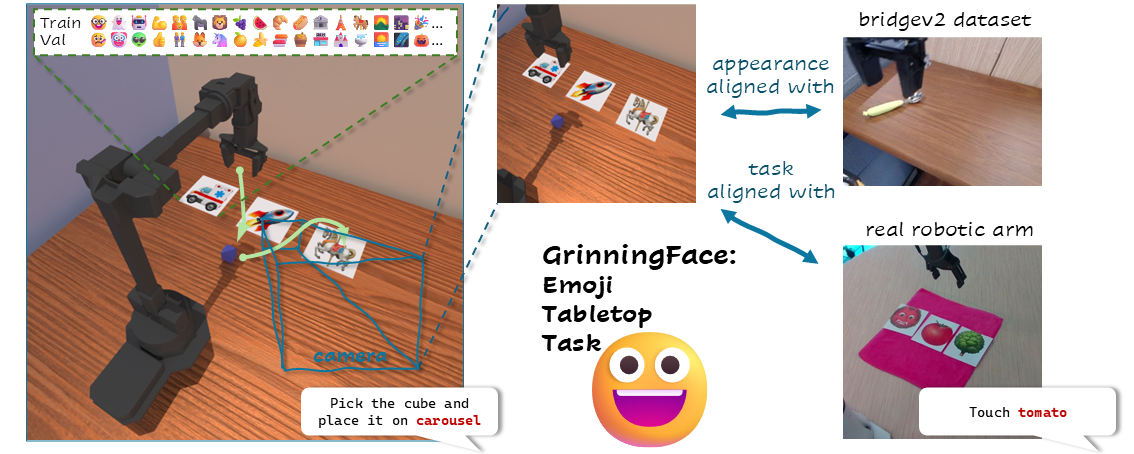}
    \caption{The GrinningFace task is designed for controlled experiments on how VLAs are pre-trained and fine-tuned to efficiently inherit the priors from VLMs. The robotic arm is asked to pick up the cube and place it on the instructed emoji. The emojis are sampled from the training set for fine-tuning and the validation set for evaluation. 
    We align the viewpoint of this task with the bridge-v2 dataset~\citep{walke2023bridgedata} to investigate the role of VLA pre-training on the models. 
    We also create a similar real robot setup to validate whether our findings in simulation holds on real robots.}
    \label{fig:grinning_face}
\end{figure*}

While the intuition that VLM-initialized VLAs can transfer prior knowledge is widely accepted, empirical validation in a controlled setting remains limited.
In this context, the key challenge in developing VLAs that excel at both robotic control and out-of-distribution generalization is \emph{catastrophic forgetting}~\citep{hancock2025actions} where the rich, general-purpose knowledge of VLMs deteriorates during further training on robotics datasets.
To address this challenge, various techniques for either pre-training or fine-tuning of VLA have been proposed, including parameter-efficient training~\citep{hancock2025actions}, freezing the VLM backbone~\citep{driess2025knowledge,nvidia2025gr00T}, co-training~\citep{zitkovich2023rt2,intelligence2025pi05}, predicting discretized action targets~\citep{driess2025knowledge,liu2025hybridvla,pertsch2025fast}, predicting latent actions~\citep{ye2024lapa,chen2025villa}, and pre-training using diverse datasets~\citep{generalist2025gen0}. 
However, these approaches are proposed in different forms and evaluated in different scenarios.
Consequently, their effectiveness remains elusive due to the absence of systematic and reproducible comparative evaluation.

The key to resolving these questions lies in constructing a set of controlled experiments with a diagnostic task that can disentangle the dual sources of VLA capabilities: 
the motor skills acquired during robotic training and the visual-semantic knowledge inherited from pre-trained VLMs. 
To this end, we introduce the emoji tabletop task, GrinningFace, where the robot is asked to place objects onto the printed emoji that corresponds to language instructions as shown in Figure \ref{fig:grinning_face}. 
We choose emojis as the proxy for two key reasons: 
First,
Emojis represent visually distinctive glyphs that are virtually absent from curated robotic manipulation datasets yet ubiquitous in the Internet-scale corpora used for VLM pre-training. 
Second, emojis span diverse categories, encompassing a wide range of concepts from everyday life. This provides a total pool of nearly 4,000 distinct symbols, a vocabulary much larger than the number of object classes in typical benchmarks. 
While our current implementation uses a subset, our framework is designed to capitalize on this scalability, holding the potential to leverage a significantly richer set of primitives.
Consequently, high task success rates in this task calls for not only motor skills learned from robotic datasets but also the activation of VLM priors through proper VLA pre-training and fine-tuning methods.

We implement this emoji tabletop task in both the ManiSkill3~\citep{tao2024maniskill3} simulation environment and real robots. 
The simulated task serves as a lightweight, reliable benchmark for comparing different techniques towards building generalizable VLAs. 
To ensure experimental rigor, we implement all techniques within a unified $\pi_0$-style codebase and controlled pre-training and fine-tuning datasets.

Our controlled experiments reveal several key insights:
1) VLM initialization, VLA pre-training, and VLA fine-tuning all contribute to robotic control, yet play different roles. 
VLM provides visual-semantic priors but not well tailored for tabletop object recognition, VLA pre-training aligns the visual-semantic knowledge to tabletop scenes that enables fast adaptation, and VLA fine-tuning specializes the model for the target task.
2) For VLA fine-tuning, full parameter tuning performs well on narrow tasks but results in catastrophic forgetting of the pre-trained priors, whereas only tuning the action head results in inadequate learning for robotic execution. 
Low rank adaptation (LoRA)~\citep{hu2022lora} strike a balance, but its adoption in fine-tuning improves the knowledge transfer from VLM to VLA inadequately.
3) Freezing VLM and using LoRA in pre-training results in good performance on GrinningFace, but they either suffer from slow adaptation in fine-tuning or inability to scale up to complex motor skills.
4) While predicting discretized action as the targets does not present positive improvement, co-training and predicting latent actions lead to better performance, which highlights them to be promising future research directions.
5)~Although our target task is simple in motor skills, VLA pre-training on more diverse datasets results in better performance, which underscores the importance of scaling up VLA pre-training.

The contributions of this work are summarized as follows.
\begin{itemize}[leftmargin=*]
    \item We introduce a minimal, reproducible benchmark to disentangle visual-semantic priors from motor skills, which can be used as a diagnostic probe to benchmark how VLMs can be adapted effectively to VLAs.
    \item We conduct systematic comparative analysis of different VLA pre-training and fine-tuning techniques within a rigorously controlled experimental framework.
    \item We provide empirically-grounded insights that inform future research directions in developing generalizable embodied agents.
\end{itemize}

%% file: sections/related_work.tex
\section{Related Work}
\label{sec:related_work}

\subsection{Improving generalization via VLM priors}

One central promise of VLA is generalizable control, which attracts much attention from researchers.
While training VLA with more robotic data~\citep{generalist2025gen0} or domain randomization~\citep{ho2021retinagan,akkaya2019solving,laskin2020reinforcement} can improve generalization, utilizing VLM is more scalable since it can endow agents with better world knowledge and visual-semantic grounding owing to the Internet-based data ingested by VLM.
Therefore, compared to previous studies that focus on different aspects of generalization~\citep{zhou2025exploring,liu2025can,pumacay2024colosseum,xing2021kitchenshift,xie2024decomposing,zheng2022vlmbench,gao2025taxonomy}, we focus on vision-semantic generalization that is one direct aspect that VLM can benefit VLA.
We believe the study on this aspect of generalization is important in unlocking the versatile generalization capabilities of VLA through inheriting VLM priors. 

One direct way to leverage VLM priors is to query the existing VLMs~\citep{ahn2022saycan,vemprala2024chatgptrobotics} or train VLM with more robotic-related data~\citep{driess2023palme}.
However, these models only target the high-level planning and cannot be used directly for low-level control.
Starting from RT-2~\citep{zitkovich2023rt2}, researchers use the pre-trained VLM weights as the initialization and train VLAs on the top of them.
However, using the pre-trained weights as the initialization does not ensure successful knowledge transfer, and catastrophic forgetting~\citep{hancock2025actions} (or spurious forgetting~\citep{zhou2025chatvla}) can be a key challenge.
For instance, \citet{chen2025internvla} finds that recent VLAs tend to overfit fine-grained motor behaviors while under-generalizing to high-level linguistic instructions.
However, existing benchmarks that conflate many factors can hardly be used for systematic diagnosis~\citep{liu2025can} and risk becoming evaluation targets that incentivize overfitting rather than genuine generalization~\citep{mees2022calvin,liu2023libero,li2024evaluating}.
In contrast, we create a diagnostic benchmark and use it for us to better understand how VLAs inherit the priors from VLMs.
This benchmark is designed with minimized execution complexity where current VLAs have made great progress, while being hard in visual-semantic generalization which calls VLAs to preserve and utilize the VLM priors effectively.
This diagnostic benchmark is motivated by the task designed in RT-2~\citep{zitkovich2023rt2} that asks the robotic arm to move the coke can to Taylor Swift.
In this paper, we extend this idea to create a diagnostic benchmark to foster the research to address the challenging catastrophic forgetting issue in inheriting priors from VLMs.

\subsection{Approaches to inherit VLM priors}
We collect a series of approaches that are promising in improving the knowledge transfer from VLMs to VLAs.

The most popular approach is co-training first adopted by \citet{zitkovich2023rt2} that trains VLA with not only robotic data but also vision-language datasets (such as VQA, image captioning, and object detecting).
Since this technique can preserve the general perception and semantic reasoning ability effectively, it becomes popular in many successors~\citep{driess2023palme,intelligence2025pi05,lee2025molmoact,zhou2025chatvla,zhou2025vision,driess2023palme,fang2025robix,chen2025internvla}.
Since co-training requires vision-language data and is computationally costly, more efficient solutions are also developed.
OpenVLA~\citep{kim2024openvla} adopts LoRA in VLA fine-tuning to achieve better performance-compute tradeoff.
Actions-as-language~\citep{hancock2025actions} uses LoRA during VLA pre-training to minimally modify the VLM backbone and avert catastrophic forgetting.
GR00T~\citep{nvidia2025gr00T} uses pre-trained and fixed VLM to provide high-level information in the hierarchical embodied AI system.

Besides, changing the training targets may also avoid the model being distracted by low-level signals.
\citet{driess2025knowledge} trains the VLM backbone with only gradients from discretized targets while freezing it from the gradients from the action expert.
\citet{liu2025hybridvla} advocates the use of additional discretized actions as the training target for better generalization.
Various forms of latent actions~\citep{schmidt2023lapo,ye2024lapa,bruce2024genie,chen2024igor} can also serve as a high-level target, keeping the VLM backbone focused on high-level motion planning.
A representative is villa-X~\citep{chen2025villa} that uses the latent action as an additional prediction target along with the canonical VLA.

While there is a broad range of techniques that may impact the knowledge transfer from VLM to VLA, ranging from the general trend that uses more diverse data to pre-train VLA~\citep{generalist2025gen0} to narrow tricks that focus on representation retentation to preserve the prior of VLM~\citep{grover2025enhancing,kachaev2025don}, our paper provides a controlled experiment setting to investigate the impact of various methods on this fundamental knowledge transfer problem.

%% file: sections/methods.tex
\section{The Emoji Table-Top Benchmark}
\label{sec:methods}

Our objective is to construct a set of controlled experiments to disentangle the capability gained from VLM priors and the capabilities obtained from training on robotic data, and evaluate them separately.
We will first describe the simulated emoji tabletop benchmark, GrinningFace, and then introduce the real robot setup to verify the findings from GrinningFace. 

\textbf{Task Definition.}
The task is to control the robotic arm to pick up a cube on the table and then place it onto the correct emoji card according to the instruction, out of three candidates.
The instruction is \texttt{ Pick the cube and place it on [desc.]} where the \texttt{[desc.]} is the relabeled description of the target emoji.\footnote{Though official emoji descriptions are provided, we relabel them using the VLM backbone and filter out the emojis with incorrect labels, to ensure that the visual-linguistic prior aligns with the VLM's understanding.}
We divide the emojis into distinct training and validation set, each containing 100 emojis.

\textbf{Data Collection.}
To collect trajectories for VLA fine-tuning, we randomly select three different emojis from the training set and place them on the table.
Then, we control the robotic arm using a rule-based program to collect the trajectory.
To avoid hard memorization, we randomize the initial positions of the cube, the emojis, and the robotic arm.
The fine-tuning dataset consists of 500 trajectories.

\textbf{Evaluation.}
To evaluate the capabilities gained from VLM priors and VLA training separately, we make the following assumption to the overall success rate (SR):
\begin{equation}
\text{overall SR} = \text{execution SR} \times \text{recognition SR}.
\end{equation}
The agent succeeds in execution if it successfully picks up the cube and place it on any of the cards, whereas the agent succeeds in recognition if it chooses the correct card out of the three candidate cards.
In practice, we record the overall SR and execution SR, and calculate recognition SR afterwards.
In this way, we can disentangle the two sources of capabilities.
The ability to recognize emojis that rarely appears in the VLA training data is inherited from the VLM backbone,
whereas the motor execution ability is gained from VLA training on robotic datasets.

To further investigate the generalization gap, we select the emojis for each trial in three protocols.
1) \texttt{ID} evaluates the model on the same emoji combinations seen in the training trajectories; 
2) \texttt{Train} evaluates the model on the emojis sampled from the training set which may form unseen combinations; 
3) \texttt{Val} evaluates the model on the emojis sampled from the validation set.
For each protocol, we evaluate the model for 100 trials which is sufficient to yield statistically meaningful results (see Appendix~\ref{app:std}).

In GrinningFace, we design the motor execution to be a simple pick-and-place and focus more on how to improve the recognition SR on \texttt{Val} which indicates the knowledge transfer from VLM to VLA in the subsequent experiments.

\textbf{Real Robot.}
For the experiment on the real robot, we design an even simpler task to evaluate more efficiently and adequately to collect statistically significant results.
The task instruction is \texttt{Touch [desc.]} that asks the agent to move the end effector on the top of the target emoji and close the gripper.
We print the emojis on cards and adopt the same training and validation set as the simulated benchmark.
We evaluate the model for 30 trials on emojis from the validation set.

%% file: sections/experiments.tex
\begin{figure*}[t]
    \centering
    \begin{subfigure}[b]{0.32\textwidth}
        \includegraphics[width=\textwidth]{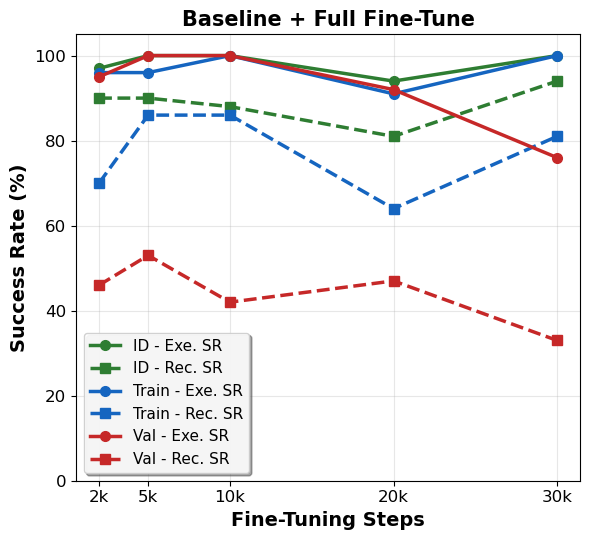}
    \end{subfigure}
    \hfill
    \begin{subfigure}[b]{0.32\textwidth}
        \includegraphics[width=\textwidth]{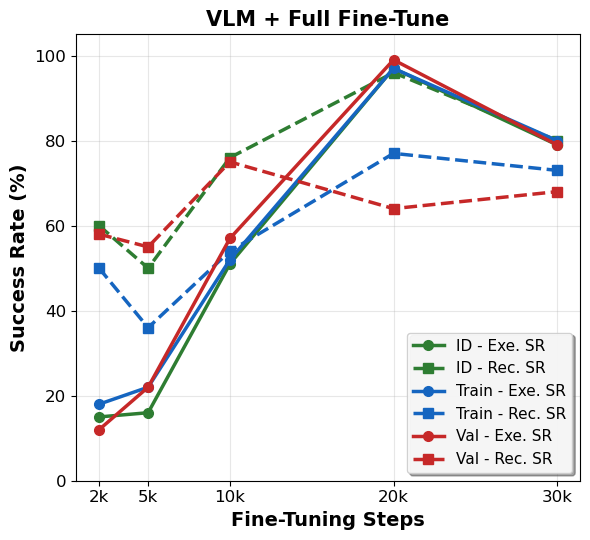}
    \end{subfigure}
    \hfill
    \begin{subfigure}[b]{0.32\textwidth}
        \includegraphics[width=\textwidth]{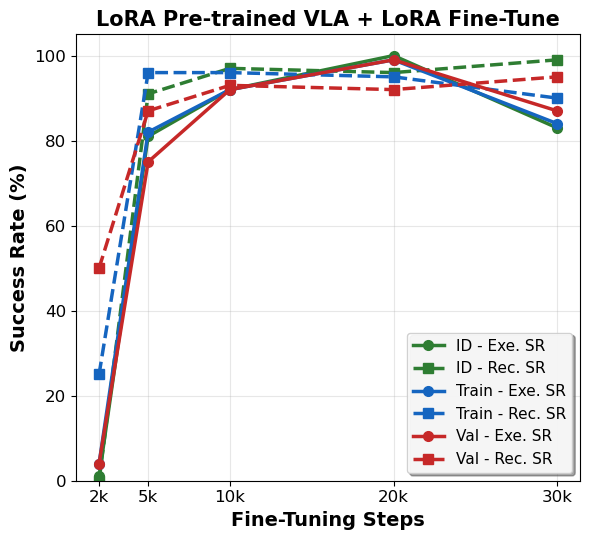}
    \end{subfigure}
    \caption{The performance w.r.t. the number of fine-tuning gradient steps on the baseline VLA with full-parameter fine-tuning (left), the VLM backbone with full-parameter fine-tuning (middle), and the VLA pre-trained using LoRA with LoRA fine-tuning (right).
    The results indicate that while VLA pre-training enables fast adaptation in fine-tuning, it degrades the priors in VLM.
    Using LoRA in pre-training and fine-tuning can well preserve the VLM priors, but it needs more fine-tuning steps to obtain even simple motor skills.
    }
    \label{fig:exp1}
\end{figure*}

\section{Experiments}
\label{sec:experiments}
We design our experiments to evaluate a series of techniques that are promising in improving the recognition SR.
For clarity, we list both the techniques of interest and brief conclusions as follows:
\begin{itemize}[leftmargin=*]
    \item \textbf{Baseline.} Our $\pi_0$-style baseline achieves high execution SR but poor recognition SR, highlighting the need for better preservation of VLM priors.
    \item \textbf{Parameter efficient fine-tuning.} 
    We use LoRA fine-tuning or only fine-tuning the action expert to preserve the knowledge in VLM. 
    While both can improve recognition SR slightly, the latter underfits in in-distribution scenarios.
    The result indicates the need to dive into the pre-training of VLA to further improve recognition SR. 
    \item \textbf{Freezing VLM backbone.} 
    We fine-tune directly from a VLM backbone or a LoRA-pre-trained VLA to maximally preserve the prior.
    This can increase recognition SR significantly, but it is not scalable since it needs much heavier fine-tuning especially when complex robotic skills are needed.
    \item \textbf{VLA co-trained with vision-language tasks.}
    We design a co-training task during VLA training, and find that co-training can lead to improved performance.
    \item \textbf{VLA trained with discretized targets.}
    We train the VLA with binned action targets, and find that using discretized targets leads to worse execution SR and recognition SR.
    \item \textbf{VLA trained with latent action targets.} 
    We train the VLA to predict the latent action along with the robot action as in villa-X~\citep{chen2025villa}, and find that this achieves better recognition SR.
    \item \textbf{VLA pre-trained on diverse data.} 
    We conduct ablation study on the VLA training data and find that pre-training VLA on diverse datasets, even if some are distinct from the target environment, can improve the performance.
\end{itemize}

We also conduct experiments on the real robot to verify the above findings, and visualize the attention map of different models to indicate the complementary roles of VLM, VLA pre-training, and VLA fine-tuning.

\begin{figure*}[t]
    \centering
    \includegraphics[width=\textwidth]{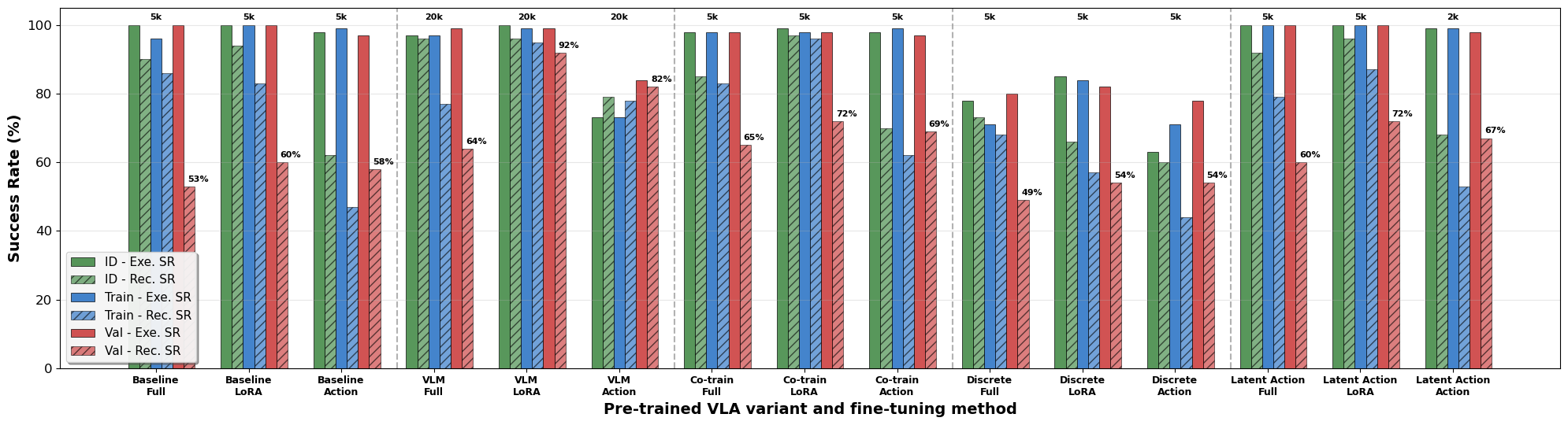}
    \caption{
    The performance of different fine-tuning methods (full parameter fine-tuning, LoRA, and only fine-tuning action expert) on different VLAs, including the baseline VLA (Baseline), VLM backbone as the VLA (VLM), the VLA co-trained with vision-language tasks (Co-train), the VLA trained with discretized targets (Discrete), the VLA trained with latent action targets (Latent Action).
    We label the execution success rate (Exe. SR) on \texttt{Val} that indicates how well VLA preserves the VLM priors.
    We also label the fine-tuning steps on which these results are gathered, and they also indicate the checkpoints that achieve the best overall success rate ($=$ Exe. SR $\times$ Rec. SR) on \texttt{Val}.
    }
    \label{fig:exp2}
\end{figure*}

\textbf{Baseline Setup.}
To compare different VLA training techniques fairly, we conduct all the experiments on the same code base, and this baseline achieves strong performance on general robotic benchmarks (see Appendix~\ref{sec:app_code_base_performance}).
Our code base is adopted from an open source implementation of $\pi_0$~\citep{black2024pi_0} that uses PaliGemma~\citep{beyer2024paligemma} as the VLM backbone and SigLIP~\citep{zhai2023sigmoid} as the vision encoder.

\textbf{Implementation details.}
By default, we train VLAs for 80 thousands (or simply 80k) gradient steps on the Open X-Embodiment~\citep{vuong2023open} magic-soup mixture containing 913k trajectories.
We fine-tune VLAs for 30k gradient steps and claim the results for the best checkpoint (regarding the overall SR on \texttt{Val}) from the checkpoints on 5k, 10k, 20k, and 30k gradient steps.
The batch size is set to 1,024 for both VLA pre-training and fine-tuning.
We pre-train and fine-tune VLAs on $8\times$ Nvidia A100, and evaluate the models on a single Nvidia A100.
We run the VLA on 100 trajectories for evaluation on each protocol, and find that this results in low standard deviation to make our results statistically significant (see Appendix~\ref{app:std}).
The code for GrinningFace is provided on \url{https://github.com/zhangchuheng123/GrinningFace}.

\textbf{Verification on the premise.}
Our basic assumptions for the GrinningFace task are as follows:
1) The motor execution ability is obtained from the VLA training on robotics dataset and fine-tuning; and 2) the emoji recognition ability sources from pre-trained VLMs. 
While the first assumption holds in obvious since VLMs are not trained to predict the robotic data, we need to assess the second assumption.

Through the following tests, we find that 1) the VLM can recognize emojis, and 2) the emojis rarely appear in VLA training data.
\begin{itemize}[leftmargin=*]
    \item First, we select three emojis and prompt the VLM to give a description for each of them. 
    (Note this may be correct even if it does not matches the official label exactly.) 
    We compute the CLIP similarity between the official label of a random target emoji and the three generated descriptions.
    We record the accuracy whether the description of the target emoji achieves the highest CLIP similarity.
    We repeat the test 100 times and find that the accuracy for \texttt{Train}, \texttt{Val}, and \texttt{Hard} are 90\%, 89\%, and 85\% respectively.
    This indicates that the VLM can recognize most of the emojis used in our experiment.
    \item The later experiment that fine-tunes the VLM backbone directly can result in over 90\% recognition SR, which also indicates that VLM can recognize emojis.
    \item In general, emoji icons do not appear in robotic datasets, but the emoji concepts may appear in data. We iterate over the task descriptions and find that less than 5\% of the emoji concepts appears in the OXE dataset.
    This indicates that the VLA must preserve the priors from VLM to achieve high recognition SR on our benchmark.
\end{itemize}

\textbf{Baseline Performance.}
We present the performance of the $\pi_0$-style baseline w.r.t. fine-tuning steps in Figure~\ref{fig:exp1} (left). 
The result shows that the model can achieve high execution SR (the solid lines) while low recognition SR (the dashed lines), especially for the emojis in the validation set,
which indicates that the baseline VLA is good at motor execution while struggles in vision-semantic recognition.
(Though the execution skill required in our benchmark is simple, Appendix~\ref{sec:app_code_base_performance} demonstrates our baseline performs well on general robotic benchmarks that does not require vision-semantic generalization.)
This suggests that further techniques are needed to help the VLA to preserve the priors from VLMs.

\textbf{Parameter efficient fine-tuning.}
We compare full parameter fine-tuning with LoRA and only fine-tuning action expert, and present the experiment results in Figure~\ref{fig:exp2} (see baseline + full/LoRA/action).
We observe that 1) all methods achieve high execution SR on different sets as the previous full fine-tuning variant; 2) LoRA fine-tuning and only fine-tuning action head both achieve slightly higher recognition than full parameter fine-tuning; and 3) only fine-tuning action head does not fit well even in in-distribution scenarios. 
Although it seems that LoRA fine-tuning can avoid knowledge forgetting, the improvement is limited and we need more VLA pre-training techniques to preserve priors from VLM.
Besides, full fine-tuning is a standard scheme in many previous work and only fine-tuning the action expert can be a good probe to compare different pre-trained VLAs.
Therefore, we will evaluate different pre-trained VLAs under all these three fine-tuning schemes hereafter.
In later experiments on different pre-trained VLAs presented in Figure~\ref{fig:exp2}, we observe that LoRA fine-tuning achieves consistently higher recognition SR on \texttt{Val}.

\begin{figure*}[t]
    \centering
    \includegraphics[width=\textwidth]{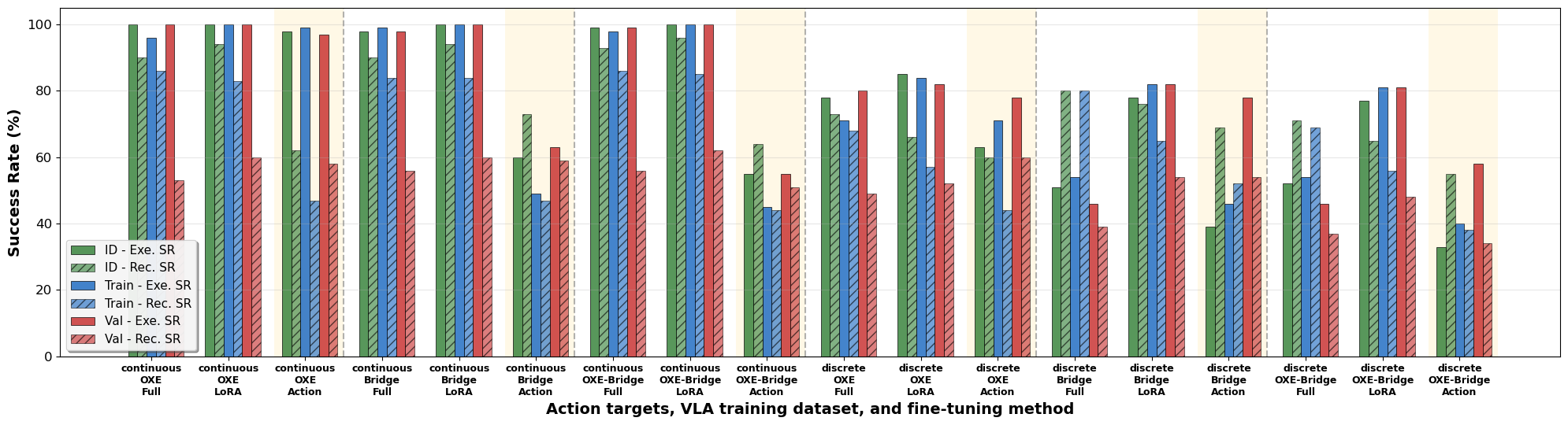}
    
    \caption{The performance w.r.t. different VLA training datasets. 
    We evaluate the VLAs pre-trained on the open-x-embodiment magic soup mixture (OXE), the bridge-v2 dataset (Bridge), and OXE excluding the Bridge dataset (OXE-Bridge) on VLAs trained with continuous targets or discretized targets.
    We highlight fine-tuning with only the action expert, which is an indicative probe to compare different pre-training datasets.
    The results indicate that training VLA on diverse dataset results in better performance.
    }
    \label{fig:exp3}
\end{figure*}

\textbf{Freezing VLM backbone.}
A direct solution to improve the recognition SR is to freeze the VLM backbone and fine-tune directly based on the VLM or a LoRA-pre-trained VLA (i.e., the VLM backbone remains frozen while VLA pre-training updates solely the LoRA adapters plus the action expert).
We present the result of directly fine-tuning VLM in Figure~\ref{fig:exp1} (middle) and Figure~\ref{fig:exp2} (VLM Full/LoRA/Action), and the result of fine-tuning the LoRA-pre-trained VLA in Figure~\ref{fig:exp1} (right).
We find that these two direct solutions result in high recognition SR, achieving over 90\% recognition SR with VLM + LoRA or LoRA-pre-trained VLA + LoRA.
However, they need 20k fine-tuning gradient steps to learn such a simple pick-and-place skill, which indicates their inability to fast adapt to more complex motor skills.
In contrast, the baseline only needs 2k to 5k gradient steps to achieve high execution SR.
Beside, our later analysis on how VLA pre-training impacts the attention map also highlights the need of sufficient VLA pre-training. 

\textbf{VLA co-trained with vision-language tasks.}
Co-training typically involves variants of vision-language tasks such as VQA and box/point/trace prediction to preserve broad VLM priors.
To make it accessible, we consider a simplified setting where the co-training task is designed to only preserve the knowledge related to emoji recognition.
We implement two variants: 
One is to co-train with the original emoji symbols and ask the VLM ``what's in the image'', and the other is to co-train with the printed emojis in the same tabletop scene and ask the VLA to recognize all the emojis.
We find that while the former one does not improve the recognition SR, the later one achieves good performance.
We present the results in Figure~\ref{fig:exp2} (Co-train Full/LoRA/Action).
The results indicate that co-training is a promising direction to enable efficient knowledge transfer, but the co-training tasks need to be carefully designed.

\textbf{VLA trained with discretized targets.}
We discretize the continuous low-dimensional action into 256 uniformly spaced bins, and overwrite the last 256 VLM tokens to represent them.
We implement two variants: 
One is to predict each dimension of the action auto-regressively as in OpenVLA~\citep{kim2024openvla}, and the other is to predict all the dimensions simultaneously based on learnable query tokens.
We find that the later one performs slightly better and report the later variant in Figure~\ref{fig:exp2} (Discrete Full/LoRA/Action).
We observe that VLA trained with discretized targets results in not only decreased execution SR but also recognition SR.
While it is predictable that using discretized actions as the target can degenerate motor control due to quantization error, we find that it still does not bring significant improvement in preserving the VLM priors.

\textbf{VLA trained with latent action targets.}
Sharing the similar motivation of co-training and prediction discretized actions, latent actions also serve as a form of high-level training target that can prevent the VLM from distracting from low-level signals. 
There are two popular ways to utilize latent actions.
One is to first pre-train VLA to predict the latent action and then fine-tune this VLA to predict the robot action (cf. LAPA~\citep{ye2024lapa}), and the other is to pre-train VLA to predict the latent action as well as the robot action and then fine-tune it to predict the robot action (cf. villa-X~\citep{chen2025villa}).
We adopt the latter one and show the result in Figure~\ref{fig:exp2} (Latent Action Full/LoRA/Action).
We observe that using latent actions as the target achieves good performance in preserving the VLM priors, which indicates that pre-training VLA with both the high-level and low-level targets may be a promising solution in knowledge transfer from VLMs to VLAs.

\textbf{Pre-training VLA on diverse data.}
With the fast accumulation of robotic data, it is also valuable to investigate whether pre-training VLA on diverse data deteriorates or activates the VLM priors.
Therefore, we conduct ablation study on pre-training datasets.
We consider the OXE magic soup dataset, the bridge v2 dataset, and OXE magic soup excluding bridge v2.
We present the results in Figure~\ref{fig:exp3}.
1)~Comparing different fine-tuning methods, we observe that different pre-training datasets influence the performance of only fine-tuning the action expert more significantly than full parameter fine-tuning or LoRA fine-tuning, indicating that fine-tuning only the action expert can serve as a good probe to compare different pre-trained VLAs.
2) Comparing Bridge and OXE-Bridge, we find that VLA performs better when the pre-training data are similar to the target scenario (noting that GrinningFace is aligned with the bridge v2 dataset visually and use the same WidowX  robot arm).
3) Comparing OXE and Bridge, we conclude that using more diverse data in VLA pre-training can help improve the performance, even if the pre-training data contains diverse datasets.

\begin{figure*}[t]
    \centering
    \includegraphics[width=\textwidth, trim={2.5cm 1.5cm 2.5cm 1.5cm}, clip]{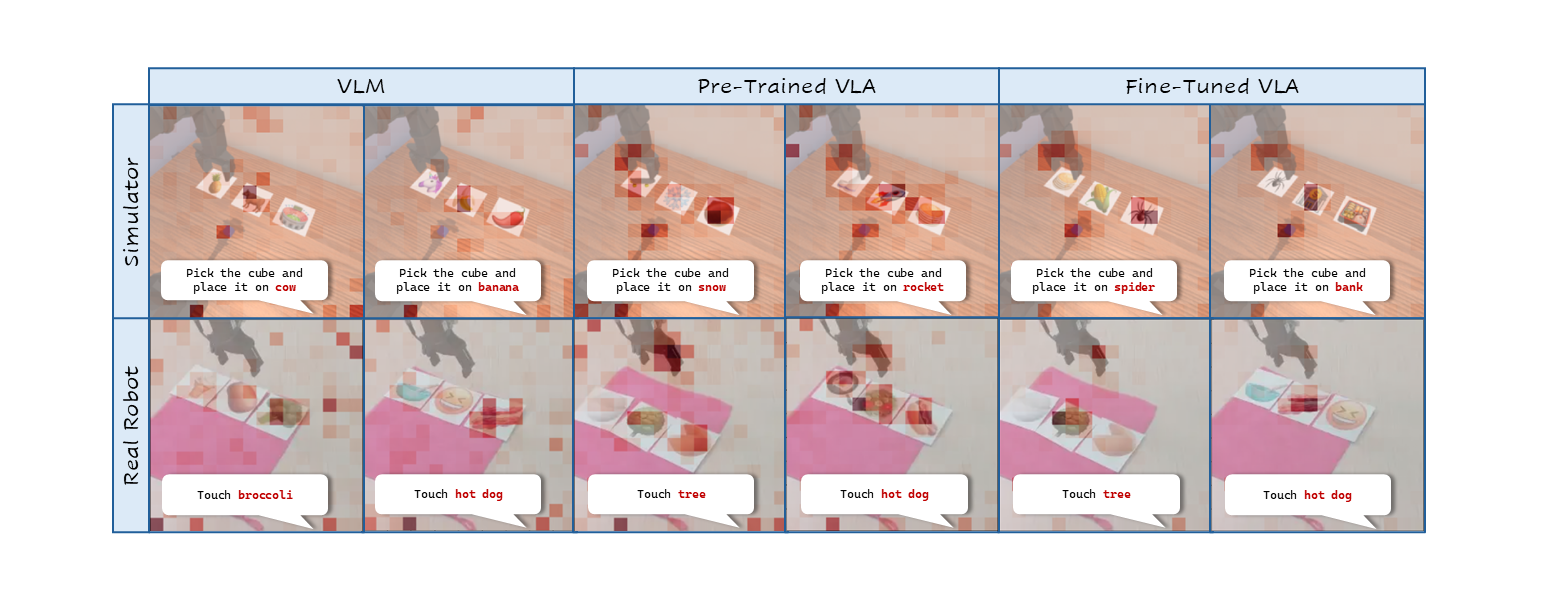}
    \caption{The attention map of different image patches to \texttt{[desc.]} (the description of the target emoji) in the task instruction for the VLM backbone, the pre-trained VLA, and the fine-tuned VLA, on both simulator and real robot.
    }
    \label{fig:attention}
\end{figure*}

\begin{table}[t]
\centering
\caption{Performance comparison of different VLA training strategies on real robot. The numbers }
\label{tab:real_robot}
\begin{tabular}{lcc}
\toprule
\textbf{Method} & \textbf{Exe. SR} & \textbf{Rec. SR} \\
\midrule
Baseline                & 30/30 & 11/30 \\
Freezing VLM            & 28/30 & 20/28 \\
Co-training             & 30/30 & 26/30 \\
Discretized action      & 30/30 & 10/30 \\
Latent action           & 30/30 & 24/30 \\
\bottomrule
\end{tabular}
\end{table}

\textbf{Experiments on real robot.}
To verify the previous findings, we conduct the emoji tabletop experiment on a real robot setup with the Realman RM75 robotic arm and the gripper from Inspire Robots.
We start from different pretrained VLAs and fine-tune them for 5k gradient steps with the same fine-tuning dataset collected by teleoperation. 
(One exception is that we fine-tune the VLM for 20k steps.)
We evaluate the success rate of different variants on the emojis from the validation set.
Execution SR indicates whether the gripper is closed when the arm is moved to any of the emojis, and recognition SR indicates that whether it chooses the correct emoji.
We evaluate each variant for 30 trials with different emoji combinations.
We present the experiment results in Table~\ref{tab:real_robot}.
The results support our main findings from simulated environment, e.g., effectiveness of co-training and predicting latent actions.

\textbf{Attention map.}
During experiment, we find that the attention map of \texttt{[desc.]} in the task instruction to the image patches is closely related to the success rate, i.e., if the \texttt{[desc.]} tokens attends much more to the correct emoji than others, the VLA will succeed in this trial with high probability.
To further analyze the important role of VLA pre-training, we visualize the attention map of the VLM backbone, the pre-trained VLA, and the fine-tuned VLA on our tasks in Figure~\ref{fig:attention}.
The models are trained following the pipeline of the baseline with full fine-tuning.
We find that 1) the VLM backbone can attend to the targeted emoji but not very focused, 2) the pre-trained VLA is more focused on meaningful objects in manipulation tasks such as the gripper, the cube, and the candidate cards, and 3) the fine-tuned VLA attend more to the correct emoji and less to the distractors (for in-distribution emojis).
While it is not surprised that the fine-tuned VLA is more skillful at the choose-one-from-three task, how VLA pre-training impacts the visual perception is interesting -- it helps the VLM to recognize the tabletop objects and therefore enables more efficient fine-tuning.


%% file: sections/conclusion.tex
\section{Conclusion}
\label{sec:conclusion}

In this work, we investigate the fundamental questions about knowledge transfer in vision-language-action models (VLAs) -- how to inherit and preserve the rich vision-semantic priors from pre-trained vision-language models (VLMs) efficiently. 
While training VLAs from VLM initialization has become the dominant paradigm with promising empirical results, 
the underlying mechanisms and effectiveness of this knowledge transfer have remained poorly understood, which hinders us to build more generalizable embodied AI systems.

To systematically investigate these questions, we introduced the emoji tabletop manipulation benchmark, GrinningFace,
a minimal yet diagnostic task that cleanly disentangles motor execution capabilities from vision-semantic recognition abilities. 
By leveraging emojis as visual targets that are ubiquitous in VLM pre-training data but absent from robotic training corpora, 
our benchmark directly measures whether VLAs successfully preserve and activate inherited VLM priors.

Through comprehensive experiments in both simulation and real robot settings, we evaluated diverse VLA pre-training and fine-tuning strategies in a controlled experiment setting. 
Our experiment results yield insightful findings, including the complementary roles of VLM and VLA pre-training/fine-tuning, the poor knowledge transfer if VLM is only used as initialization, the effectiveness of co-training and predicting latent actions, etc. 

Our results highlight a critical gap: 
Current methods remain limited in seamlessly integrating VLM priors into VLA systems; without this capability, a VLA will not only struggle on real‑world tasks demanding open‑ended world knowledge, but will even fail on our simple benchmark.
While some of the techniques studied in this paper offer improvements, 
substantial room exists for developing more effective strategies to preserve and activate inherited knowledge during VLA training.
We hope our benchmark and the evaluation framework will serve as a valuable diagnostic tool for the community, 
enabling principled comparison of future techniques aimed at building truly generalizable embodied agents.

%% file: sections/app_benchmark_codebase.tex
\section{Performance of the code base}
\label{sec:app_code_base_performance}

To indicate that our experiments are conducted based on a high-performing code base, we evaluate the performance of the VLA trained using this code on two popular robotic benchmarks, SIMPLER and LIBERO.
We present the results in \ref{tab:benchmark_comparison} and observe that the performance of our baseline is comparable to other strong baselines.

\begin{table*}[h]
\centering
\small
\begin{tabular}{lcccc}
\toprule
\textbf{Benchmark} & \textbf{Octo} & \textbf{OpenVLA} & $\pi_0$ & \textbf{Baseline} \\
\midrule
SIMPLER (google) & 15 & 33 & 59 & 37 \\
SIMPLER (widowx) & 16 & 1 & 27 & 49 \\
LIBERO-spatial & 79 & 85 & 97 & 82 \\
LIBERO-object & 86 & 88 & 99 & 92 \\
LIBERO-goal & 85 & 79 & 96 & 89 \\
LIBERO-long & 51 & 54 & 85 & 83 \\
\midrule
Average & 55 & 57 & 77 & 72 \\
\bottomrule
\end{tabular}
\caption{Performance comparison on robotics benchmarks (success rate \%). }
\label{tab:benchmark_comparison}
\end{table*}

%% file: sections/app_std.tex
\section{The standard deviation of our evaluation}
\label{app:std}

To make sure that our evaluation protocol results in statistically meaningful results, we repeat the evaluation for our baseline VLA with full fine-tuning 5 times with different seeds (each with 100 trials) and record the performance.
We find that this evaluation protocol results in low standard deviation.

\begin{table}[h]
\centering
\caption{Performance across different random seeds. All values are percentages (\%).}
\label{tab:seeds}
\begin{tabular}{ccccccc}
\toprule
\multirow{2}{*}{\textbf{Seed}} & \multicolumn{2}{c}{\textbf{ID}} & \multicolumn{2}{c}{\textbf{Train}} & \multicolumn{2}{c}{\textbf{Val}} \\
\cmidrule(lr){2-3} \cmidrule(lr){4-5} \cmidrule(lr){6-7}
& \textbf{Exe. SR} & \textbf{Rec. SR} & \textbf{Exe. SR} & \textbf{Rec. SR} & \textbf{Exe. SR} & \textbf{Rec. SR} \\
\midrule
1 & 94 & 81 & 91 & 64 & 92 & 47 \\
2 & 89 & 79 & 92 & 77 & 95 & 43 \\
3 & 89 & 81 & 93 & 70 & 93 & 44 \\
4 & 91 & 78 & 92 & 74 & 94 & 47 \\
5 & 92 & 77 & 95 & 74 & 94 & 45 \\
\midrule
Mean & 91.0 & 79.2 & 92.6 & 71.8 & 93.6 & 45.2 \\
Std & 1.9 & 1.5 & 1.4 & 4.6 & 1.0 & 1.5 \\
\bottomrule
\end{tabular}
\end{table}